\documentclass[10pt,twocolumn,letterpaper]{article}

\usepackage{iccv}
\usepackage{times}
\usepackage{epsfig}
\usepackage{graphicx}
\usepackage{amsmath}
\usepackage{amssymb}

\newcounter{lenumerate}

\renewenvironment{itemize}{\begin{list} {\labelitemi}
{\setlength{\parsep}{0.1ex} \setlength{\topsep}{0.6ex}
\setlength{\partopsep}{0.1ex } \setlength{\itemsep}{0.2ex}
\setlength{\leftmargin}{2ex} }} {\end{list}}

\usepackage[pagebackref=true,breaklinks=true,letterpaper=true,colorlinks,bookmarks=false]{hyperref}

\iccvfinalcopy 

\def\iccvPaperID{****} 

\ificcvfinal\fi
\begin{document}

\title{Visual Understanding of Multiple Attributes Learning Model of X-Ray Scattering Images
}

\author{Xinyi Huang\\
{\it \small Kent State University}\\
\and
Suphanut Jamonnak\\
{\it \small Kent State University}\\
\and
Ye Zhao\\
{\it \small Kent State University}\\
\and
Boyu Wang\\
{\it \small Stony Brook University}\\
\and
Minh Hoai\\
{\it \small Stony Brook University}\\
\and
Kevin Yager\\
{\it \small Brookhaven National Laboratory}\\
\and
Wei Xu\\
{\it \small Brookhaven National Laboratory}\\
}

\maketitle

\begin{abstract}
   This extended abstract presents a visualization system, which is designed for domain scientists to visually understand their deep learning model of extracting multiple attributes in x-ray scattering images. The system focuses on studying the model behaviors related to multiple structural attributes. It allows users to explore the images in the feature space, the classification output of different attributes, with respect to the actual attributes labelled by domain scientists. Abundant interactions allow users to flexibly select instance images, their clusters, and compare them visually in details. Two preliminary case studies demonstrate its functionalities and usefulness.
\end{abstract}

\section{Introduction}

X-ray scattering helps scientists discover molecular and nano level physical structures of materials such as nano-particles, protein, lithographic gratings, polymer films, and so on. The technique is widely used in biomedical, material, and physical applications by analyzing structural patterns in the x-ray scattering images \cite{Yager2014}. X-ray equipment can generate up to 1 million images per day which impose heavy burden in post image analysis. A variety of image analysis methods are applied to x-ray scattering data. Recently, deep learning models are employed in classifying and annotating multiple image attributes from experimental or synthetic images, which were shown to outperform previously published methods \cite{Wang2017,Guan2018}.

As most deep learning paradigms, these methods are not easily understood by material, physical, and biomedical scientists. The lack of proper explanations and absence of control of the decisions would make the models less trustworthy. While considerable effort has been made to make deep learning interpretable and controllable by humans \cite{Choo2018}, the existing techniques are not specifically designed for the scientific image classification models of x-ray scattering images, which requires extra consideration in finding 
\begin{itemize}
    \item How the learning models perform for a diverse set of overlapped attributes with high variation?
    \item How the co-existence of attributes in x-ray images may affect the classification results? 
\end{itemize}
Please note that in general, these questions may also be applicable to other deep learning models for multiple object detection and segmentation, such as (fast-, faster-) R-CNN \cite{Zhao2019}, YOLO \cite{Redmon2017}, and SegNet \cite{Badrinarayanan2017}. Unfortunately, few existing visualization tools \cite{Choo2018} are focused on visually analyzing the learned results of multiple attributes, objects, or segments with these models.

In response, we develop a visual analysis system for users to interactively study the model predictions with respect to the multiple attributes with x-ray scattering images. The system has several features:
\begin{itemize}
    \item Image instances are projected and visualized in three vector spaces: actual labeling space from domain scientists, feature space extracted by a residual network, and prediction space of the model output. Users can interactively explore the instances in these spaces by performing visual comparison, outlier detection and drill-down study of images.
    \item An ``attribute-flower'' visualization (based on Astor charts) is used to represent the scientific images in an embedded space where each attribute is represented as a ``petal''. The attribute recognition results, compared with the ground truth labels, are easily depicted to discover false positive (FP), false negative (FN), true negative (TN), and true positive (TP) predictions.
    \item The visual system integrates multiple coordinated views, together with user interactions, to facilitate iterative exploration and comparison.
\end{itemize}
The system alleviates domain scientists' efforts in understanding the performance of deep learning models for x-scattering images. They can identify outliers or spurious data clusters, and from this either improving the training data or the learned model. This extended abstract shows the system design and preliminary case studies.

\section{Related Work}
The x-ray scattering data is analyzed to recognize the image attributes such as ring, halo, diffuse scattering, etc. \cite{Kiapour2014}. Recently, deep learning models are employed for the x-scattering data \cite{Sullivan2019, Park2017,Liu2019}. Wang et al. \cite{Wang2017} apply Convolutional Neural Networks (CNN) over both experimental and synthetic images to detect important attributes. Guan et al. \cite{Guan2018} further develop a DVFB-CNN model which combines Fourier-Bessel transform (FBT) within CNN model. 

Choo et al. \cite{Choo2018} categorize explainable deep learning into three major directions: model understanding, model debugging, and model refinement. Computational approaches discover importance scores of the input features contributing to the prediction results. Perturbation methods \cite{Alvarez-Melis2017}, saliency based methods \cite{Selvaraju2016}, LIME \cite{Ribeiro2016}, and influence functions \cite{Koh2017} are proposed for the purposes. These methods have not been specifically designed for x-ray scattering images.

Interactive visualization tools are developed in providing in-depth understanding of how deep learning models work. Tools such as Tensorflow \cite{Wongsuphasawat2018} Playground, Tensor Board, and ConvNetJS \cite{ConvNetJS} allow users to visualize and interact with the activation maps and network structures, together with line graphs and histograms of characteristic statistics. DeepVis \cite{Yosinski2015} shows that optimizing synthetic images with better natural image priors produces more recognizable visualizations. CNNVis \cite{Liu2017} system helps designers in their understanding and diagnosis of CNNs by exploring the learned representations in the graph layout of the neural networks. ActiVis \cite{Kahng2018} integrates embedding view with multiple coordinates views for visual model exploration. Embedding Projector \cite{EmbeddingProjector} visualizes input images in a 2D or 3D embedding space (by PCA or t-SNE), to reveal the relationship among these instances. Our approach also works in the embedding spaces by visualizing the image instances with a new attribute-flower visualization. The system allows users to analyze the physical attributes recognition results with respect to the learning models. This feature makes the system different from most existing methods working in embedding spaces, in which image, audio and national language datasets are projected and visualized by linking the final decision with the origin images/text data elements, but not the individual attributes (objects). 

\section{Learning Model of Multiple Attributes}
Automatic attribute recognition in x-ray scanned image data is a challenging problem due to the high variation in the same classes. The same structural attributes (patterns) can be of great variety in their appearances. On the other hand, a diverse set of characteristic attributes, from the type of measurement, e.g. ‘small-angle x-ray scattering (SAXS)’ or ‘wide-angle x-ray scattering (WAXS)’, to instrumental information, e.g. ‘linear beam stop’ or ‘beam off image’, to appearance-based scattering features, e.g., ‘halo’ or ‘ring’, may co-occur and overlap in the scattering images. Scientists want to seek help from deep learning models to identify these features in given images. In our prototype, we utilize an open x-ray scattering dataset \cite{Yager2017} and the ResNet model in \cite{Wang2017} for our design of the visual analytics techniques.

\noindent \textbf{Experimental and Synthetic Images}
We retrieved about 1,000 x-ray scattering images from the dataset \cite{Yager2017} to show the visual system functions. They include different types of images including semiconductors, nano-particles, polymer, lithographic gratings, and so on. The attributes in these images are either labelled by domain experts or synthetically generated by a simulation algorithm \cite{Wang2017}. Each image thus has an actual attribute vector (\textbf{ACT vector}) consisting of 17 Boolean (0 or 1) values to show if the image has one of the 17 attributes. 
\begin{figure*}[t]
 \centering
 \includegraphics[width=\textwidth]{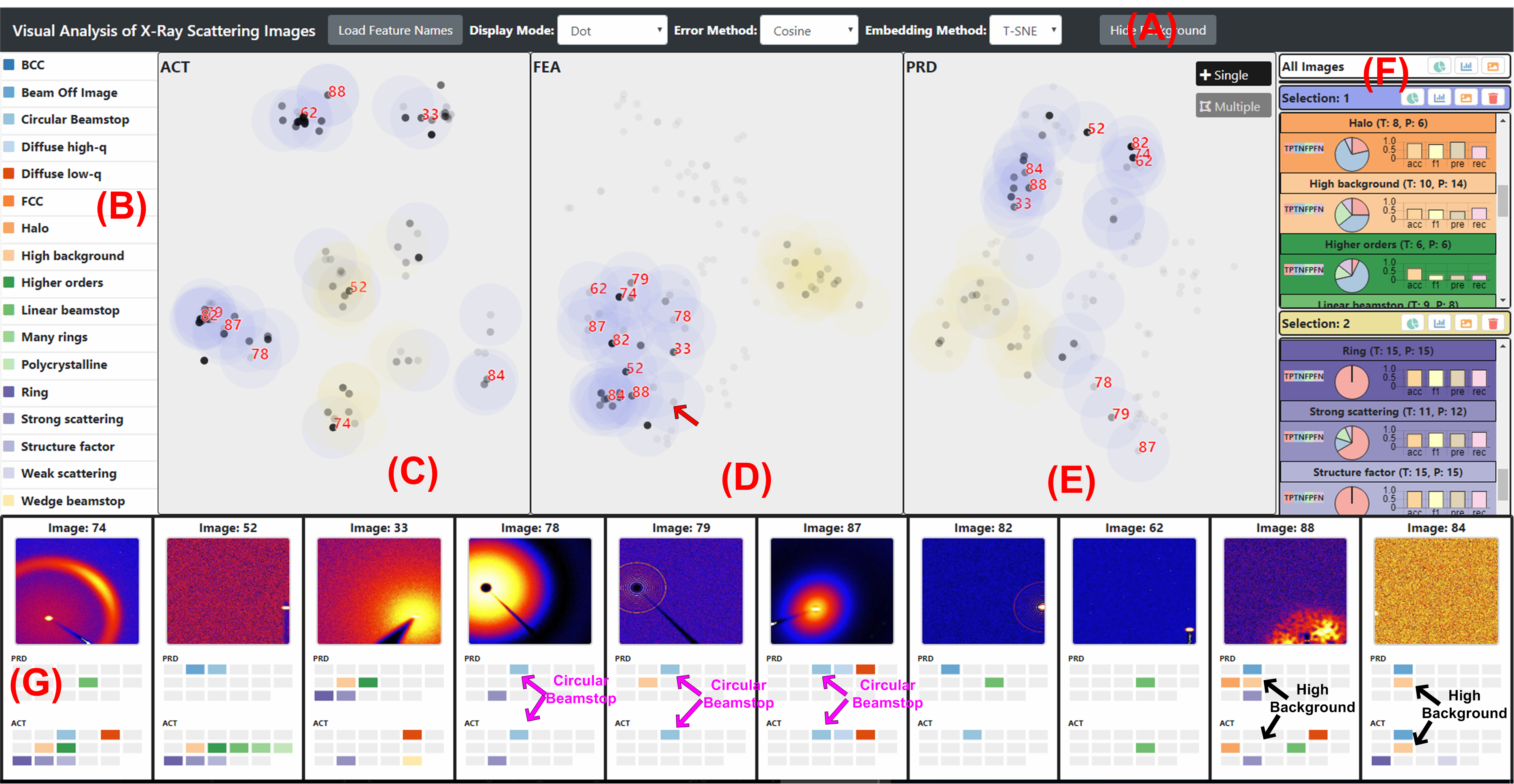}
    \caption{\it The System Interface: (A) Control panel. (B) Attribute filters. (C) Embedded view in ACT space. (D) Embedded view in FEA space. (E) Embedded view in PRD space. (F) Selection manager with group characteristics. (G) Detail image view. }\vspace{-15pt} 
    \label{fig_interface} 
\end{figure*}

\noindent \textbf{ResNet Machine Learning Model}
A 50-layer Residual Network (ResNet) proposed in \cite{He2016} was trained for the 17 attributes by more than 100,000 x-ray images. A feature vector (\textbf{FEA vector}) with 2048 dimensions was learned by the model for each image. The fully connected layers generated final output of a prediction vector (\textbf{PRD vector}) for the 17 attributes. The deep learning model reported the mean average precision (mAP) about 77\% \cite{Wang2017}. 

\noindent \textbf{Embedded Spaces and Model Metrics}
Three vectors of the x-ray images are used in the t-SNE algorithm to get the corresponding embedding coordinates. The ACT vector has 17 dimensions with each being the Boolean label corresponding to one attribute. The PRD vector with the same size 17 is obtained from the output of the trained ResNet model with fully connected layers for classification \cite{Wang2017}. Each PRD element is the prediction probability (0.0 to 1.0) for each attribute with a cut-off value of 0.5 for final decision. The FEA vector with the size of 2048 dimensions contains the activation values of the last feature extraction layer in ResNet \cite{He2016}. The characteristics of an image are supposed to be well extracted and representative by its FEA vector, though it is not directly interpretive. Comparing the images in these embedded spaces can reflect the performance of the ResNet for x-ray scattering images.
 
Additionally, four metrics are adopted to evaluate the output of the model as following formulas: Accuracy = $\frac{TP + TN}{TP + TN + FP + FN}$, Precision = $\frac{TP}{TP+FP}$, Recall = $\frac{TP}{TP+FN}$, 
and F1 = $2 \times \frac{Precision\times Recall}{Precision+ Recall}$.
The accuracy is affected by the sum of TP and TN. The precision (recall) suffers only when FN (FP) is large. F1 score instead makes both FN and FP from precision and recall in balance. They all provide the characteristic information of the model.  
In our system, these measures are visualized to evaluate the trained model such as attributes, and user selected image groups.

\section{Visual System}
\noindent \textbf{Tasks:} The learning model is usually evaluated by the metric of precision and recall for different attributes, and the mAP (mean average precision) for the images. However, it can hardly help scientists explain and make sense of how the model performs on the scientific images with respect to the structural patterns. Therefore, we identify the tasks for which visual analytics techniques can contribute to enhance the understanding:
 
    \noindent \textbf{T1.} Overview the image dataset within the spaces of ACT, PRD and FEA vectors. It can help users find and compare the instances, so as to understand how the images are modelled in the feature space, prediction space, with respect to the real labels. The precision, recall, accuracy and f1 values are visualized to show the general characteristic of the model. \\
    \noindent \textbf{T2.} Show the image instances to identify false positive (FP), true positive (TP), false negative (FN), and true negative (TN) attributes, find outliers and clusters, and understand how these measures are related to the learning model.\\
    \noindent \textbf{T3.} Select and explore image groups, the attributes of interests and their combinations, to perform the first two tasks;\\
    \noindent \textbf{T4.} Interactively visualize individual images and compare them for the model prediction performance.


\begin{figure*}[t]
 \centering
 \includegraphics[width=0.95\textwidth]{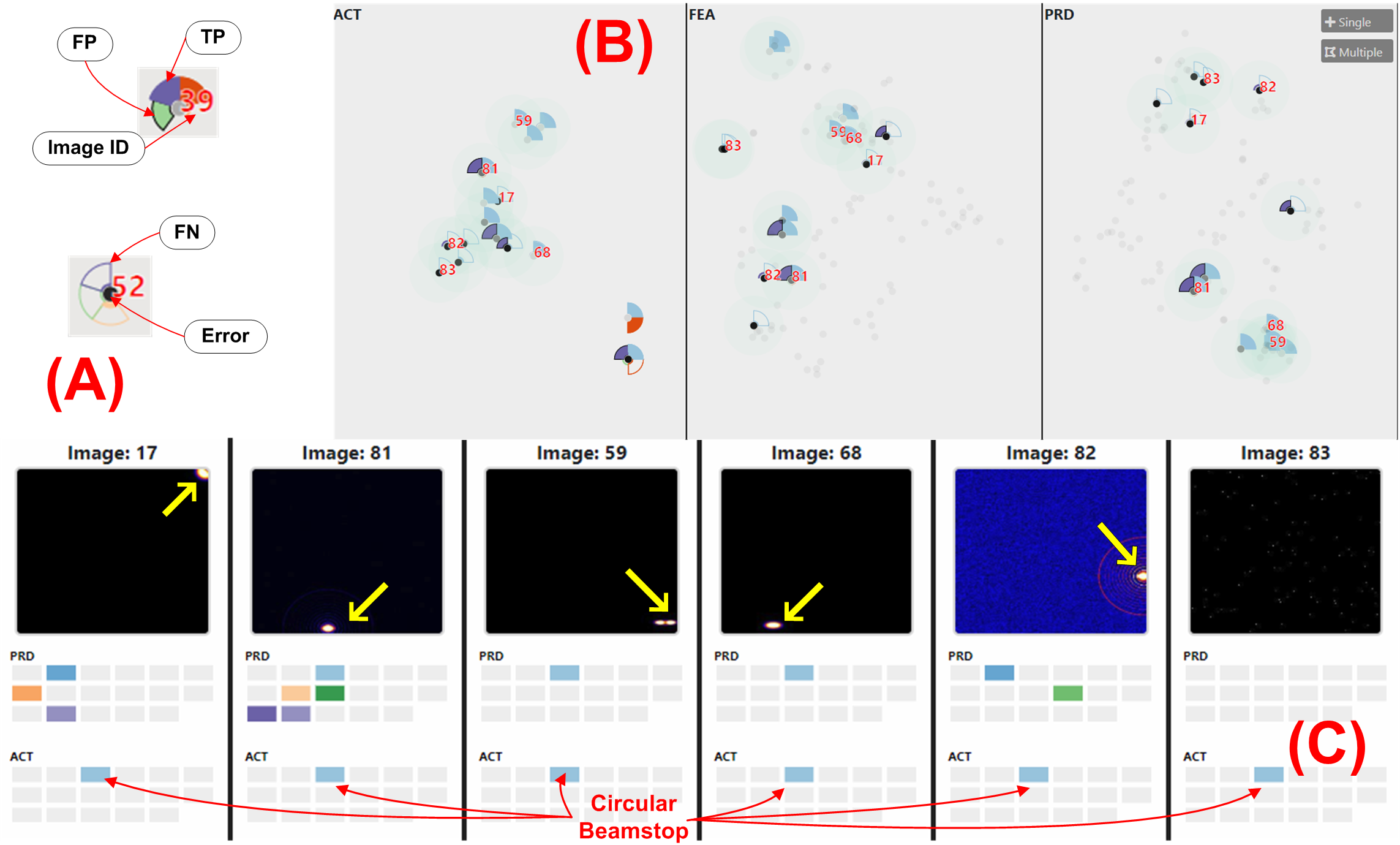}
    \caption{\it (A) The attribute flower visualizations of instance images in the joint mode. (B) Study of attribute flowers in the embedded views. (C) Drill down study of individual images selected in (B). These images are combined here to save paper space.}\vspace{-15pt} 
    \label{fig_Case2} 
\end{figure*}

\noindent \textbf{Visual Interface:}
Fig. \ref{fig_interface} shows the visual system interface designed based on the tasks. It displays the instance images in the coordinated views (C)-(E) of three different embedded spaces (ACT, FEA, PRD). Users can interactively select (with zooming and panning) groups of instances in either embedded view to find their distributions in the other views. They can also drill-down for detail study of images in the image view (G). Each image in the embedded space can be displayed as dots (C)-(E), or as attribute flowers (see Fig. \ref{fig_Case2}) displaying actual values or model predictions of different attributes. Users can filter the visualizations with single or multiple attributes (B). A selection includes multiple instances, whose group characteristics of TP, TN, FP, FN are shown in the selection manager (F), together with the precision, recall and F1 metrics. 

In particular, users can choose to use the attribute flower visualization to show each attribute as a ``petal'' (Fig. \ref{fig_Case2}(A)). Users can choose different modes to show only the ACT values or the PRD values in the flower. Moreover, a joint mode can show the status of the attributes: the center dot shows the error distance (cosine or Euclidean) between the ACT vector and PRD vector, and each ``petal'' is filled in the corresponding color of one selected attribute if this image has this attribute in its ACT vector. The petal has a black border if this attribute does not exist in the ACT vector but appears in its PRD vector. Therefore, FN, FP and TP attributes can be easily discerned in the embedded views. A supplemental video ({\it http://vis.cs.kent.edu/xscattering/}) shows the use of the interactive visual system.

\section{Case Studies}
\noindent \textbf{Case 1:}
As shown in Fig. \ref{fig_interface}(D), a set of images (with purple halo) are selected which are close in the FEA space. That is, the ``hidden'' features identified by the 50-layer network in ResNet are similar in these images. In the PRD view (Fig. \ref{fig_interface}(E)) these images are separated by the fully-connected layers in the ResNet. For example, Images 78,79,87 are close but they are apart from the other images. They also exist in a cluster in ACT view (Fig. \ref{fig_interface}(C)). By studying them in the detail view (Fig. \ref{fig_interface}(G)), they have a specific attribute \textit{Circular Beamstop} which are correctly identified (shown by the purple arrows). Meanwhile, Images 84 and 88 are close in the PRD view but far away from each other in the ACT view (Fig. \ref{fig_interface}(C)). From Fig. \ref{fig_interface}(G), both of them are identified to have an attribute \textit{High Background} referring to``noisy'' background (shown with black arrows). However, in the ACT values Image 84 is labelled to have this attribute but Image 88 is not. Image 88 does have some kind of noise background from the image while the expert did not label it. This may help scientists to understand how the learning model can identify noises. They may revise the label, or they may use more training data to let the model know what types of noises are to be identified or ignored.

\noindent \textbf{Case 2:}
By using attribute flowers to visualize images in ACT view with 4 selected attributes of interest, a cluster of images are selected as shown in a zoom-in view (Fig. \ref{fig_Case2}(B)). They are related to the attribute \textit{Circular Beamstop}. Images 17, 82 and 83 are FN in this attribute and Images 81,68,59 are TP. It can be seen that FEA does not separate them directly, while PRD plays a good role in separating them. When studying details in Fig. \ref{fig_Case2}(C), it can be seen that the model performs good for Images 59 and 68 but Image 81 discovers several additional attributes, such as \textit{Ring}. When Image 81 is fully enlarged, human eyes can identify several dim rings around the spot (yellow arrow). It shows the model can subtly identify very weak ring effects which are not even labelled by scientists. However, Images 82 and 17 are not satisfied where the \textit{Circular Beamstops} are not discovered, possibly linked to the fact that the spots (yellow arrows) are on the edge and corner of the images (this is further justified by more similar images). This variance makes the learning model fail to recognize it, which demands a new measure of the learning process and training datasets. Moreover, Image 83 has an incorrect label in the raw data.


\section{Conclusion and Future Work}
We present a visual system for understanding the learning model of x-scattering images with multiple attributes. The system allows users to visually discover the embedded distributions of feature vectors, predictions, and actual labels of these images. User interactions are supported to compare selected instance images and study their prediction results related to the attributes. Next, we will extend the work for model debugging and refinement, by taking neurons and different network layers into account, so as to study how they relate to the prediction of multiple attributes. 


{\small
\bibliographystyle{ieee}
\bibliography{XRayAI}
}

\end{document}